\documentclass[10pt,twocolumn,letterpaper]{article}

\usepackage[pagenumbers]{cvpr} 
\usepackage{graphicx}
\usepackage{amsmath}
\usepackage{amssymb}
\usepackage{booktabs}
\usepackage{multirow}

\usepackage{pifont}

\usepackage[pagebackref,breaklinks,colorlinks]{hyperref}

\usepackage[capitalize]{cleveref}
\crefname{section}{Sec.}{Secs.}
\Crefname{section}{Section}{Sections}
\Crefname{table}{Table}{Tables}
\crefname{table}{Tab.}{Tabs.}

\usepackage[dvipsnames,table,xcdraw]{xcolor}
\hypersetup{
    colorlinks=true,
    linkcolor=red,
    citecolor=RoyalBlue,
    filecolor=magenta,      
    urlcolor=cyan,
}

\usepackage[misc]{ifsym}
\newcommand\blfootnote[1]{%
\begingroup
\renewcommand\thefootnote{}\footnote{#1}%
\addtocounter{footnote}{-1}%
\endgroup
}

\begin{document}

\title{Temporal Perceiving Video-Language Pre-training }

\author{Fan Ma$^{1,2*}$, Xiaojie Jin$^{1*}$\textsuperscript{\tiny \Letter}, Jingjia Huang$^{1}$, Heng Wang$^{1}$, Linchao Zhu$^{2}$, Jiashi Feng$^{1}$\textsuperscript{\tiny \Letter}, Yi Yang$^{2}$\\
$^{1}$ByteDance Inc. $^{2}$Zhejiang University\\
}
\maketitle

\blfootnote{$^{*}$Equal contribution. Fan Ma did the work while interning at ByteDance Inc. \newline \hspace*{1.9em}{\scriptsize \Letter} Corresponding author: Xiaojie Jin$<$\url{jinxiaojie@bytedance} \\ \url{.com}$>$, Jiashi Feng$<$\url{jshfeng@bytedance.com}$>$.} 
\begin{abstract}
      Video-Language Pre-training models have recently significantly improved various multi-modal downstream tasks. 
    Previous dominant works mainly adopt contrastive learning to achieve \emph{global} feature alignment across modalities. 
    However, the local associations between videos and texts are not modeled, restricting the pre-training models' generality, especially for tasks requiring the temporal video boundary for certain query texts. 
    This work introduces a novel text-video localization pre-text task to enable fine-grained temporal and semantic alignment such that the trained model can accurately perceive temporal boundaries in videos given the text description.
    Specifically, text-video localization consists of moment retrieval, which predicts start and end boundaries in videos given the text description, and text localization which matches the subset of texts with the video features.
    To produce temporal boundaries, frame features in several videos are manually merged into a long video sequence that interacts with a text sequence. 
     With the localization task, our method connects the fine-grained frame representations with the word representations and implicitly distinguishes representations of different instances in the single modality.
    Notably, comprehensive experimental results show that our method significantly improves the state-of-the-art performance on various benchmarks, covering text-to-video retrieval, video question answering, video captioning, temporal action localization and temporal moment retrieval. Codes will be released.

\end{abstract}

\section{Introduction}
\label{sec:intro}

\begin{figure}[!t]
    \centering
    \includegraphics[width=\linewidth]{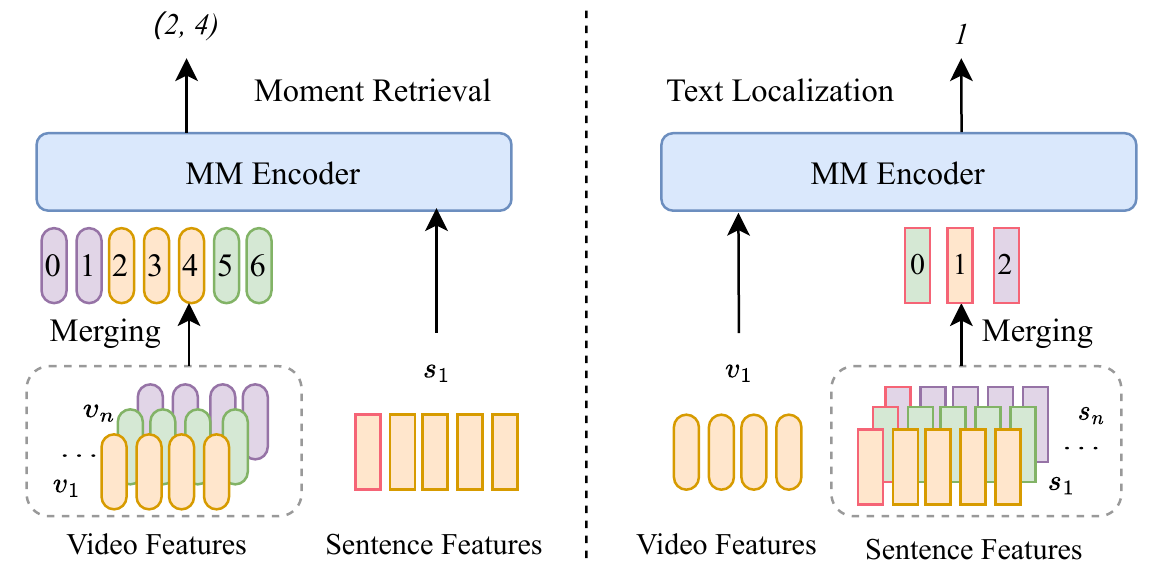}
    \caption{\textbf{Temporal Perceiving Video-Language (TemPVL) pre-training}. The left part is the moment localization given a sentence query, and the right part is the text localization task for the video query. The frame and text features are represented with the ellipse and rectangle, respectively. The paired video and text features are marked with the same color, and the [CLS] token in every sentence is red outlined.
    }
    \label{fig:intro} 
\end{figure}

Video-language pre-training that learns generic representations from large-scale multi-modality data has been popular in the past two years~\cite{clip2021, xu2021videoclip, Fu2021VIOLETE, wang2022all, huang2022clover, Cao2022LocVTPVP}.
The pre-trained  models demonstrate excellent multi-tasks capabilities, covering video question answering (QA)~\cite{xu2016msr, xu2017video}, text-to-video retrieval~\cite{anne2017localizing, maharaj2017dataset}, and video
captioning~\cite{Lin2022SwinBERTET, xu2017video}, under various settings of zero-shot, few-shot or transfer learning~\cite{wang2022all, Cao2022LocVTPVP}. 
By aligning representations between large-scale video-text pairs, the pre-trained video-language models have achieved encouraging performance on various applications~\cite{clip2021, huang2022clover, xu2021videoclip}. 

Contrastive learning is widely used in the video-language pre-training by globally aligning the video with text representations~\cite{clip2021, clipbert2020}.
However, the video often contains irrelevant frames to the text, and the global contrastive learning dismisses the fine-grained alignment between frames and texts. 
For instance, a party video with caption ``the person is cutting cake'' may contain frames where children run around the table. 
This would not only limit the matching performance between videos and texts, but also lead the model to fail to learn distinguished visual features for temporal localization. 
The fine-grained temporal-aware alignment between relevant texts and video frames is thus essential to learn generic multi-modal representations.

A few works~\cite{Fu2021VIOLETE, ge2022bridging, wang2022all} recently adopted masked language modeling (MLM) to enhance the fine-grained interactions by predicting the masked element with unmasked visual and text features.
Albeit MLM works well on multiple reasoning tasks such as video QA and captioning, the fine-grained alignment along the temporal dimension  is not guaranteed. 
LocVTP~\cite{Cao2022LocVTPVP} manages to form the fine-grained contrastive loss by splitting the video into several clips and extracting phrases from sentences, but the objective is constructed on the pseudo alignment since the temporal annotations are not available.
In practice, the temporal annotations are either unavailable in current short video-text pairs (WebVID~\cite{Bain21}), or heavily noisy in long videos (HowTo100M~\cite{miech19howto100m}). The reliable video-text pairs with accurate annotations are still missed for fine-grained temporal alignment modeling.

To augment the temporal modeling ability of video-language models for better perceiving fine-grained interaction between videos and texts, we introduce a novel text-video localization pre-training task where temporal annotations are no longer required. The proposed task consists of two objectives as shown in~\cref{fig:intro}, the moment localization with language and the text localization with video. 

Specifically, we follow the mainstream settings~\cite{huang2022clover, ge2022bridging} to adopt dual encoders for encoding video and language inputs separately and use a multi-modal encoder to fuse both visual and text features.
For the moment localization, the frame features of several videos are merged into a single long video sequence where the temporal position of each video can be inferred from the merging strategy and  used for pre-training the model.
The merged frame features interact with text tokens in the multi-modal encoder to enhance alignment between the fine-grained frame and text features.
Similarly, we merge multiple text tokens for text localization. The matched text positions to a video will be predicted to correlate frame features with all text features.
With text-video localization, fine-grained frame-word alignment is well established and temporal context modeling is implicitly encoded.

Extensive experimental results on several downstream tasks also demonstrate the superiority of our proposed text-video localization pre-training task. Our method improves zero-shot text-to-video retrieval performance by 3.3$\%$ on DiDeMo and acquires 2.6$\%$ performance gain on the moment retrieval task. 
In summary, our contributions are three-fold.

\begin{itemize}
    \item We present a novel video-text localization task for video-language pre-training where temporal modeling across multi-modalities is well designed and fine-grained interaction between visual and language signals is encouraged. 
    \item With the text-video localization pre-training task, the generalization capability of the pre-training model is consistently improved across different tasks and backbones. 
    
    \item Comprehensive experiments on five downstream tasks demonstrate the superiority of our method. In addition, off-the-shelf models in temporal action localization and moment retrieval tasks, can further boost performance by using our extracted video features.
\end{itemize}

\section{Related Work}
\label{sec:rela}
\subsection{Video-Language Pre-training}
Large-scale multi-modal data has been leveraged to build pre-training video-language (VidL) models Pre-training recently.  
Pre-trained models exhibit surprising generalization capacities when fine-tuning on a series of popular downstream video-language tasks, including text-to-video retrieval~\cite{xu2016msr, anne2017localizing}, video question answering~\cite{xu2016msr, xu2017video}, and video captioning~\cite{Lin2022SwinBERTET, xu2016msr}. 
Contrastive learning is widely used in video-language pre-training to project videos and texts into the identical feature space~\cite{miech2020end, xu2021videoclip}. 
Contrastive learning only coarsely aligns the representations between videos and text descriptions.
To enable multi-modal interactions, several models, such as  VideoBERT~\cite{sun2019videobert}, HERO~\cite{li2020hero},
ActBERT~\cite{zhu2020actbert}, ClipBERT~\cite{clipbert2020}, MERLOT~\cite{zellers2021merlot}, SwinBert~\cite{Lin2022SwinBERTET}, VIOLET\cite{Fu2021VIOLETE}, All-in-One~\cite{wang2022all}, adopt popular masked language modeling (MLM) also to predict masked signals. 
However, the fine-grained alignment is still not achieved in these works, limiting the model generalization capacity.
LocVTP~\cite{Cao2022LocVTPVP} manages to build fine-grained alignment by introducing the clip-phrase contrastive objective. However, the objective is based on pseudo supervision where the matching between clips and phrases is not granted. 
In this work, we introduce a novel text-video localization task to encourage fine-grained video and text feature alignment without any annotations, achieving significant improvement on multiple video-text downstream tasks.

\subsection{Video Temporal Modeling}
Temporal modeling is a critical yet challenging topic in video understanding, containing action recognition and localization tasks. 
Prominent ideas including sparse sampling~\cite{TSN2016ECCV, feichtenhofer2019slowfast}, spatial-temporal operations~\cite{tran2014c3d, bertasius2021space} are introduced for temporal modeling in both convolution and Transformer
architectures~\cite{tran2014c3d, bertasius2021space}.
To enhance the temporal modeling for obtaining better video representations, TSP~\cite{Alwassel2021TSPTP} trains video encoders to be temporally sensitive by predicting clips inside or outside the action where temporal annotations are required in training datasets.
All-in-One~\cite{wang2022all} manages to enhance temporal interaction by rolling the video features in temporal dimension. 
LocVTP~\cite{Cao2022LocVTPVP} explicitly models the clip-word matching from the video language pairs based on pseudo supervision. 
In this work, we construct a long video sequence in the training batch and feed them to the multi-modal encoder, where frame features interact with temporal contexts and text features to predict accurate temporal boundaries.

\section{Method}
\label{sec:method}

\subsection{Preliminary}

In this section, we present TemPVL, a new text-video localization pre-text task for pre-training.
We follow the prominent architecture~\cite{xu2021videoclip, huang2022clover} that uses two encoders for extracting video and text features separately and one multi-modal encoder for both visual and text features. 
Given the video input, the visual encoder $E_{v}$ outputs the video features $ \mathbf{f}_{v} \in \mathbb{R}^{C_{v} \times T \times h \times w} $ where $C_{v}$ denotes the feature channel, and $h$ and $w$ denote the down-scaled spatial resolution.
We adopt BERT-base architecture for both the multi-modal and text encoders. The input text is first tokenized into a token sequence, where two special tokens [CLS] and [SEP] are inserted at the beginning and the end respectively. The text encoder $E_{s}$ outputs the token features $\{\mathbf{f}_{w}^{i}\ \in \mathbb{R}^{C_{w}}\}_{i=0}^{L+1}$ where $C_{w}$ is the feature channel and there are $L$ tokens in each sentence.
To fuse both the visual and text features, the video feature is first pooled along the spatial dimension into frame tokens. All the frame and text tokens are then projected into the common embedding space to form the concatenated  multi-modal input $\{\mathbf{f}_{m}^{i} \in \mathbb{R}^{C}\}_{i=0}^{T+L+1}  $, where the first $T$ tokens come from the video frames. Every frame token interacts with all the frame and word tokens in the multi-modal encoder to learn unified representations for both visual and language inputs. 

Previous pre-training methods use contrastive loss on the paired video and text embedding to align the cross-modal representations. However, the contrastive learning only encourages the global video-text matching, lacking correspondence between the individual frames and words. On the one hand, the temporal modeling on video frames is not well established with coarse text-video contrastive learning. On the other hand, the representations are not well aligned, limiting the performance on downstream extensions, such as the text-to-video retrieval.
To achieve the fine-grained multi-modal alignment, we introduce a novel text-video localization pre-training task.

\begin{figure}
  \centering
    \includegraphics[width=\linewidth]{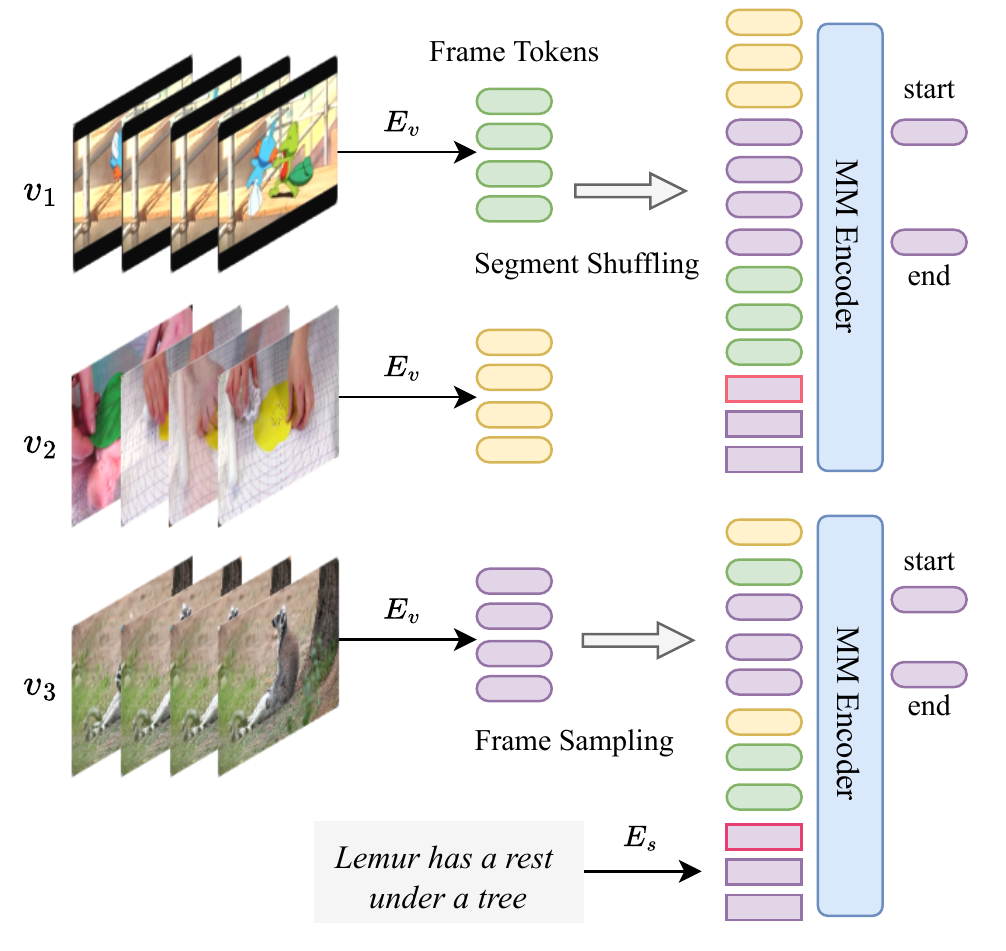}
    \caption{\textbf{Moment retrieval with different video merging strategies}. The frame tokens of videos are first extracted through the visual encoder $E_{v}$. A long frame sequence is then obtained by merging all frame tokens. The frame tokens are concatenated with the text tokens to form the input for the multi-modal encoder. The multi-modal encoder encourages the interactions between all frame and text features by predicting the start and end 
   position of the clip that aligns with the text description.}
    \label{fig:mergevideo}
  
\end{figure}

\subsection{Text-Video Localization}
In the multi-modal encoder, the frame tokens interact with text tokens to update representations. Previous methods adopt masked language modeling or video-text matching to enhance the interaction. However, masked language modeling mainly benefits the reasoning task, such as video question answering and video captioning, failing to supervise the fine-grained alignment. The video-text matching uses positive and negative video-text pairs and does the binary classification on the text [CLS] token, which is also the rough alignment. 
In this section, we present text-video localization to enable fine-grained alignment between different modalities. 

Specifically, our TemPVL predicts the temporal boundary from video tokens for the text description, and produces the localization from the word tokens given the visual input.
For most video-text pre-training datasets, short video-text pairs are common and the temporal annotations are usually unavailable that indicate which clips in videos are aligned with text descriptions. 
To enable the temporal alignment, we form the long video sequences by merging frame features of different videos in the training batch and get the long paragraph descriptions by merging word tokens of different sentences. 
Next, we present each localization task in detail.

\subsubsection{Moment Retrieval with Language}

The moment retrieval is to temporally localize the clip in the video related to the text description as shown in \cref{fig:mergevideo}. During the pre-training, the short videos in a training batch are used to constitute a long video sequence. For each sentence, the start and end frames of the moment that match the texts are predicted.

\noindent\textbf{Video Merging.}
As each video contains multiple frames, we constitute a long video sequence by combining frames in different videos. 
Instead of directly merging video frames, we first extract all frame features via the visual encoder and combine the videos via concatenating the frame features. 
Given the video feature, we use spatial mean pooling followed by a linear projection layer to get frame tokens $\mathbf{t}_{v} \in \mathbb{R}^{C\times T}$. 
The $T$ is usually small as pre-training models use a few frames in each video. 
To construct a long sequence, we adopt two ways to merge videos as shown in \cref{fig:mergevideo}. 
The first way is to concatenate frame features of different videos by shuffling video segments. 
Suppose we have $B$ videos in a training batch, a sequence with $BT$ frame tokens $\mathbf{t}_{mv} = [\mathbf{t}_{v_1}, ..., \mathbf{t}_{v_B}] \in \mathbb{R}^{BT \times C}$ is formed where the temporal order of frames in the video is retained and the video order is randomly permuted. 
The merged frame tokens are added with frame position embeddings and then concatenated with word tokens in one sentence to constitute the multi-modal input tokens $\mathbf{t}_{m} \in
 \mathbb{R}^{C \times (BT+L+2)}$. 
We also use frame sampling strategy to generate long video sequences. Specifically, we define the total number of frame tokens as $K$ and the number of positive frame tokens as $K_p$. We sample $K_p$ frame tokens from the video that correlates to the text and $K-K_{p}$ background tokens from the rest of videos. The positive frame tokens is randomly inserted in the background tokens. 
For both merging strategies, the temporal boundary for the combined text tokens can be easy inferred  and denoted as $(st_{v},ed_{v})$.

\noindent\textbf{Boundary Prediction.}
In multi-modal encoder, all tokens interact with each other so the frame tokens can encode text information to improve the representation and vice versa.
To strengthen the interactions between different modalities, we introduce a localization task where the temporal boundaries for the merged text are required to be predicted.
We apply two linear layers with a norm layer to output frame tokens for producing localization predictions $\mathbf{r}_{vl} \in \mathbb{R}^{BT \times 2}$. The localization objective is written as:
\begin{equation}
    \mathcal{L}_{vl} = -\log\mathrm{softmax}(\mathbf{r}_{vl}^{0})^{st_{v}} -\log\mathrm{softmax}(\mathbf{r}_{vl}^{1})^{ed_{v}},  
\end{equation}
where $\mathbf{r}_{vl}^{0} \in \mathbb{R}^{BT}$ is the start logit prediction and $\mathbf{r}_{vl}^{1}$ is the end logit prediction. We use softmax operation to get the start and end probability of all frame tokens for the text. The $st_{v}$ and $ed_{v}$ are the ground-truth start and end indices.  
Different from the regression loss in many temporal action localization, the classification loss is used in our moment localization task to supervise the learning process. As only one matched video in the merged frame tokens, the classification loss is simple yet effective to encourage the coherence between visual and text features. 
By converging the predictions with the start and end positions, the frame features absorb temporal contexts and align with the text representation.

\begin{figure}
  \centering
    \includegraphics[width=\linewidth]{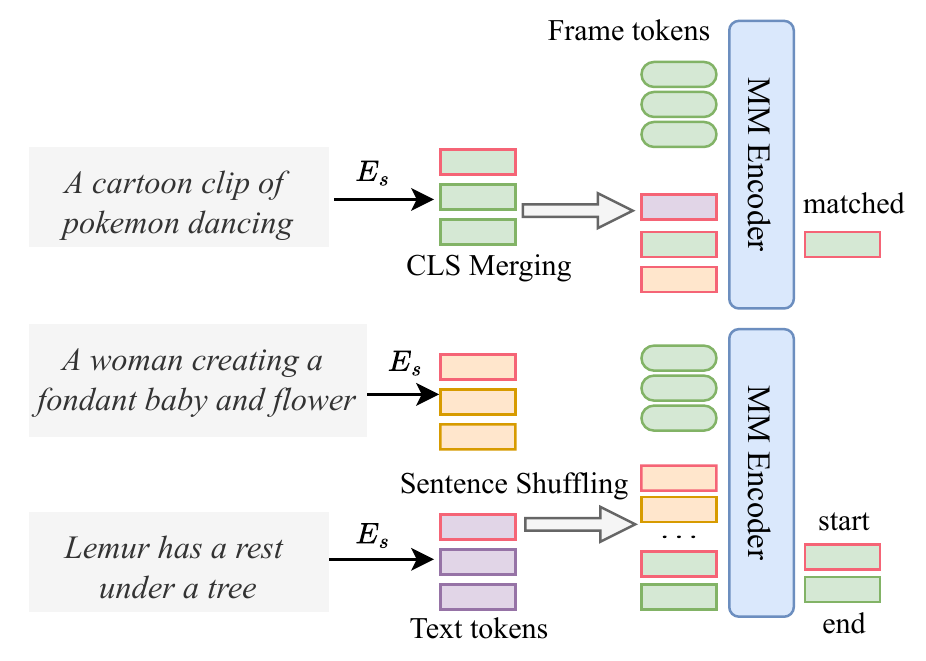}
    \caption{\textbf{Text localization for video with different merging strategies}. We merge word tokens in two ways to form a long text sequence. All the merged word tokens are concatenated with one video tokens to feed the multi-modal encoder. The text localization is to predict the matched position from the output text tokens for the fused frame tokens. }
    \label{fig:text_merge}
  
\end{figure}

\subsubsection{Text Localization with Video}
The text localization is to localize the boundary in text tokens for a video clip. The text could contain both relevant and irrelevant parts for the video clip. Similar to moment retrieval, we merge word tokens from several training sentences and concatenate them with frame tokens in one video.

\noindent\textbf{Text Merging.}
As shown in \cref{fig:text_merge}, we can also use the shuffling strategy to combine different sentence tokens. 
 All [CLS] and [SEP] tokens in every sentence are also merged in the shuffling strategy.
The sampling merging in moment retrieval is not feasible for the text tokens as the same words or phrases could be contained in different sentences, and the semantic information is changed if only a few words are sampled.  
In addition, we propose to merge texts by only combining the [CLS] token in every sentence. For one video and $B$ merged text tokens, we formulate the multi-modal input tokens as $\mathbf{t}_{m} \in \mathbb{R}^{C \times (T+B)}$.

\noindent\textbf{Classification.}
We adopt the same start and end prediction loss when the shuffling strategy is adopted on the text tokens. For the second merging strategy, we only predict the matched [CLS] token index since only one token is used for each sentence.
\begin{equation}
    \mathcal{L}_{tl} = -\log\mathrm{softmax}(\mathbf{r}_{tl})^{m_{t}},  
\end{equation}
where $\mathbf{r}_{tl} \in \mathbb{R}^{B}$ is the logit prediction and $m_{t}$ is the matched text index for the video.

\subsection{Pre-training Objectives}

The text-video contrastive is to coarsely project video and text into the common feature space, while the text-video localization is to enhance fine-grained visual language interaction for video-text alignment. With the mask language modeling, our pre-training objective is formed via:

\begin{equation}
    \mathcal{L} = \mathcal{L}_{vtc} + \alpha\mathcal{L}_{mlm} + \beta\mathcal{L}_{vtl},
\end{equation}
where $\mathcal{L}_{vtl} = \mathcal{L}_{vl} + \mathcal{L}_{tl}$ denotes the text-video localization task. The $\alpha$ and $\beta$ are the hyper-parameters to balance each pre-training tasks, which are set to 1 in our experiment. The $\mathcal{L}_{vtc}$ and $\alpha\mathcal{L}_{mlm}$ denote  contrastive learning and masked language modeling.


\begin{table*}
  \centering
  \resizebox{1.0\textwidth}{!}{
  \begin{tabular}{l|ccc|ccc|ccc|ccc}
    \toprule
    	\multicolumn{1}{l|}{\multirow{2}{*}{Method}} & \multicolumn{3}{c|}{MSR-VTT} & \multicolumn{3}{c|}{DiDeMo} & \multicolumn{3}{c|}{MSVD} & \multicolumn{3}{c}{LSMDC} \\
    \multicolumn{1}{c|}{}  & Recall@1 & Recall@5 & Recall@10 & Recall@1 & Recall@5 & Recall@10  & Recall@1 & Recall@5 & Recall@10  & Recall@1 & Recall@5 & Recall@10  \\
    \midrule 
    \multicolumn{13}{c}{Zero-shot} \\ 
    \midrule
    NoiseEst~\cite{amrani2021noise} &8.0 &21.3 &29.3 &- & - & - &13.7 &35.7 &47.7 &4.2 &11.6 &17.1 \\ 
    SupportSet~\cite{patrick2020support} &12.7 &27.5 &36.2 &- & - &- &21.4 &46.2 &57.7 &- & - &- \\
    VideoCLIP~\cite{xu2021videoclip} &10.4 &22.2 &30.0 &16.6 &46.9 &- & - & - &- & - & - &- \\
    Frozen~\cite{bain2021frozen} &18.7 &39.5 &51.6 &21.1 &46.0 &56.2 &33.7 &64.7 &76.3 &9.3 &22.0 &30.1\\
    Clover~\cite{huang2022clover} & 25.8 & 49.6 & 60.1  & 28.0 & 53.5 & 65.1  & & &  &13.8 &28.1 &38.3 \\
    VIOLET~\cite{Fu2021VIOLETE} & 25.9 & 49.5 & 59.7 &23.5 &49.8 &59.8 & - & - &- &- &- &- \\
    LocVTP~\cite{Cao2022LocVTPVP} & 22.1 & 48.0 & 55.3 &- &- &- &- &- &- &- &- &- \\
    \midrule
    TemPVL (Ours) &\textbf{27.2} &\textbf{50.1} &\textbf{60.3} &\textbf{31.3} &\textbf{56.4} &\textbf{67.6}  & \textbf{41.6} & \textbf{72.3} & \textbf{80.6} &\textbf{14.9} &\textbf{31.2} &\textbf{39.5} \\ 
    \midrule
    
    \multicolumn{13}{c}{Fine-tune} \\ 
    \midrule
    NoiseEst~\cite{amrani2021noise} &17.4 &41.6 &53.6 &- & - &- &20.3 &49.0 &63.3 &6.4 &19.8 &28.4 \\
    SupportSet~\cite{patrick2020support} &30.1 &58.5 &69.3 &- & - &- &28.4 &60.0 &72.9 &- & - &- \\
    VideoCLIP~\cite{xu2021videoclip} &30.9 &55.4 &66.8 & - & - &- & - & - &- & - & - &-\\
    Frozen~\cite{bain2021frozen} &31.0 &59.5 &70.5 &31.0 &59.8 &72.4 &45.6 &79.8 &\textbf{88.2} &15.0 &30.8 &39.8 \\
    Clover~\cite{huang2022clover} & 38.6 & 67.4 & 76.2 & 45.1 & 74.3 & 82.2 & & & & 22.7 & 42.0 & 52.6 \\
    Lavender~\cite{li2022lavender} & 37.8 & 63.8 & 75.0 & 47.4 & 74.7 & 82.4 & 46.3 & 76.9 & 86.0 & 22.2 & \textbf{43.8} & \textbf{53.5} \\
    VIOLET~\cite{Fu2021VIOLETE} &34.5 &63.0 &73.4 &32.6 &62.8 &74.7 & - & - & - &16.1 &36.6 &41.2\\
    LocVTP~\cite{Cao2022LocVTPVP} & 36.5 & 64.3 & 76.8 & - & - & - & - & - & - & - & - & - \\
    All-in-one~\cite{wang2022all} &37.9 &68.1 &77.1 &32.7 &61.4 &73.5 & - & - & - & - & - & - \\
    \midrule
    TemPVL (Ours) &\textbf{41.0} &\textbf{68.2} &\textbf{77.7} &\textbf{48.6} &\textbf{76.1} &\textbf{85.4} & \textbf{47.8} & \textbf{79.7} & 87.2  &\textbf{23.2} &42.2 & 51.3  \\
    \bottomrule
  \end{tabular}}
  \caption{Text-to-video retrieval comparison on MSR-VTT, DiDeMo, MSVD, and LSMDC under the zero-shot and fine-tune setups. Higher Recall@k indicate better performance. The best performance is masked in bold under each setting.}
  \label{tab:retrieval_comp}
\end{table*}

\section{Experiments}
\label{sec:exp}

\subsection{Datasets and Downstream Tasks }
\noindent\textbf{Pre-training datasets.}
Following recent work~\cite{huang2022clover}, we jointly pre-train our TemPVL on the WebVid~\cite{bain2021frozen} with 2.5M video-text pairs and the Google Conceptual Captions
(CC3M)~\cite{Sharma2018ConceptualCA} with about 3M image-text pairs. The static image is treated as the video with only single frame during the pre-training.

\noindent\textbf{Downstream tasks.}
We evaluate our method on five popular downstream tasks.
(1) \textbf{Text-to-video retrieval} on four datasets: MSR-VTT~\cite{xu2016msr}, DiDeMo~\cite{anne2017localizing}, MSVD~\cite{xu2017video} and LSMDC~\cite{maharaj2017dataset}. This task evaluates how well the text representations align with the video features. 
(2) \textbf{Video question answering} task on MSR-VTT~\cite{xu2016msr} and MSVD~\cite{xu2017video}. The open-ended setting is adopted in QA to evaluate the reasoning ability of the pre-training models. 
(3) \textbf{Video captioning} that requires understanding the action and event in the video on MSR-VTT~\cite{xu2016msr} and MSVD~\cite{xu2017video}.
(4) \textbf{Temporal action localization} on THUMOS~\cite{THUMOS14}. Perceiving temporal context and discriminative frame features are significant in this task. 
(5) \textbf{Video moment retrieval} is similar to temporal action localization where the text query is engaged to localize temporal boundaries in videos. We conduct experiments on DiDeMo~\cite{anne2017localizing} to testify the pre-training models.

\subsection{Implementation Details}
We adopt VideoSwin~\cite{Liu2022VideoST} as the video encoder with pre-trained weights on the Kinetics-400 dataset~\cite{Kay2017TheKH}, and pre-trained BERT-base model as the text encoder. 
The multi-modal encoder is initialized from the last three layers of the pre-trained BERT-base model.
All modules are end-to-end tuned during both pre-training and fine-tuning. We pre-train our model for 40
epochs, using a batch size of 2048 on 64 NVIDIA A100 GPUs. We use AdamW~\cite{Loshchilov2019DecoupledWD} optimizer with a weight decay 0.005 and betas (0.9, 0.98). 
The learning rate is first set to 5e-5 and then
decays by 10 times following a cosine annealing decay schedule. All video frames are resized to 224$\times$224, and 8 frames are randomly sampled in a video while the temporal order is preserved. During pre-training, all words in the sentence is random masked with 15\% probability to enable the mask language modeling in both normal and causal attentions. 
For the retrieval task, we only fine-tune the uni-modal encoders with the contrastive learning. 
For both video QA and video captioning tasks, we adopt the casual mask in both text and multi-modal encoders to generate both answers and descriptions. 
For both temporal action localization and moment retrieval with language tasks, we use the pre-trained visual encoder to extract video features first and adopt the off-the-shelf algorithms to train corresponding models with our extracted features.

\subsection{Comparison to Prior Arts}

\subsubsection{Text-to-Video Retrieval}

\cref{tab:retrieval_comp} illustrates the text-to-video retrieval results on MSR-VTT~\cite{xu2016msr}, DiDeMo~\cite{anne2017localizing}, MSVD~\cite{xu2017video} and LSMDC~\cite{maharaj2017dataset} datasets under zero-shot and fine-tuning settings.
Clover~\cite{huang2022clover} and LocVTP~\cite{Cao2022LocVTPVP} are  also pre-trained on WebVid~\cite{bain2021frozen}+CC3M~\cite{Sharma2018ConceptualCA}) where Clover uses the ranking loss to sort the alignment between different modalities, and LocVTP introduces fine-grained contrastive learning on the pseudo frame-phrase matching predictions.
Our proposed method significantly outperforms the previous approaches among all the datasets. 
Notably, the performance improvement with zero-shot evaluation demonstrates the stronger generalization ability of our method.
Our TempVL achieves the highest recall on four datasets under the zero-shot setting. 
In detail, our method outperforms Clover~\cite{huang2022clover} by 1.4$\%$ on MSR-VTT, 3.3$\%$ on DiDeMo, 1.1$\%$ on LSMDC in Recall@1. 
Moreover, our proposed method surpasses VIOLET by a large margin on both MSR-VTT and DiDeMo, even though VIOLET is pre-trained with more text-video pairs (\emph{i.e.,} WebVid~\cite{bain2021frozen}+CC3M~\cite{Sharma2018ConceptualCA}+YTT180M~\cite{Zellers2021MERLOTMN}).

When fine-tuned on the four datasets, TemPVL also shows superiority over the compared methods. Our method outperforms the compared methods across all the metrics on MSR-VTT and DiDeMo with a clear improvement. Compared to videos in MSR-VTT, videos in DiDeMo contain more frames and diverse scenes. The noticeable improvement on the DiDeMo also suggest that our method  better matches long videos with texts.  Compared to LocVTP~\cite{Cao2022LocVTPVP} that leverages pseudo fine-grained alignment information, our model with the text-video localization pre-training task achieves much higher results under both zero-shot and fine-tune settings.

\subsubsection{Video Question Answering}

\cref{tab:vqa_comp} shows the results on two open-ended video question answering datasets. We compare our method with several methods, including JustAsk~\cite{Yang2021JustAL}, ALPRO~\cite{Li2022AlignAP}, VIOLET~\cite{Fu2021VIOLETE}, All-in-one~\cite{wang2022all}, Clover~\cite{huang2022clover} and Lavender~\cite{li2022lavender}. Different from All-in-one and Clover that use classification loss for the open-ended QA, we use the generative way to produce answers where the categories are not limited.
Our method outperforms all other methods on the MSR-VTT dataset and surpasses Lavender by 0.4$\%$.

\begin{table}
  \centering
  \begin{tabular}{l|c|c}
    \toprule
    	\multicolumn{1}{l|}{\multirow{2}{*}{Method}} & \multicolumn{1}{c|}{MSR-VTT} & \multicolumn{1}{c}{MSVD}  \\
    	 &Accuracy  &Accuracy \\    
    \midrule 
    JuskAsk~\cite{Yang2021JustAL} &41.5 &46.3 \\
    ALPRO~\cite{Li2022AlignAP} &42.1 &45.9 \\
    VIOLET~\cite{Fu2021VIOLETE} &43.9 &47.9 \\
    All-in-one~\cite{wang2022all} &44.3 &47.9 \\
    Clover~\cite{huang2022clover} & 43.9 & 51.9 \\ 
    Lavender~\cite{li2022lavender} & 44.2 & \textbf{55.4} \\
    \midrule    TemPVL (Ours) &\textbf{44.7} &53.9 \\
    \bottomrule
  \end{tabular}
  \caption{Video question answering comparison on MSR-VTT and MSVD under the open-ended setting. We report the accuracy and the highest performance is masked in bold.}
  \label{tab:vqa_comp}
\end{table}

\subsubsection{Video Captioning}

We present the comparison on video captioning task in \cref{tab:caption_comp}. 
For this task, the causal mask is used in both text and multi-modal encoders and 60$\%$ of words are masked during fine-tuning. We compare our method with four pre-training models, including DECEMBERT~\cite{tang2021decembert}, SwinBERT~\cite{Lin2022SwinBERTET}, MV-GPT~\cite{Seo2022EndtoendGP} and Lavender~\cite{li2022lavender}. The results show that our model achieves the highest CIDEr score on both MSR-VTT and MSVD datasets, demonstrating the capability of the proposed TemPVL.

\begin{table}
  \centering
  \begin{tabular}{l|c|c}
    \toprule
    \multicolumn{1}{l|}{\multirow{2}{*}{Method}} & \multicolumn{1}{c|}{MSR-VTT} & \multicolumn{1}{c}{MSVD}  \\
    	 &CIDEr  &CIDEr \\
    \midrule 
    DECEMBERT ~\cite{tang2021decembert} & 52.3 & - \\ 
    SwinBERT~\cite{Lin2022SwinBERTET}  &53.8  &120.6 \\
    MV-GPT~\cite{Seo2022EndtoendGP}    &60.0  &- \\
    Lavender~\cite{huang2022clover}  & 58.0  & 142.9  \\
    \midrule
    TemPVL (Ours) & \textbf{61.9} & \textbf{148.2} \\
    \bottomrule
  \end{tabular}
  \caption{Video captioning comparison on MSR-VTT and MSVD under the open-ended setting. We report the CIDEr score and and the highest score is masked in bold.}
  \label{tab:caption_comp}
  
\vspace{-10pt}
\end{table}

\subsubsection{Temporal Action Localization}

We go a step further to evaluate the effectiveness of our proposed method on temporal action localization task. We follow the previous setting~\cite{Alwassel2021TSPTP} to only extract video features after pre-training and use the GTAD~\cite{xu2020gtad} to train the temporal action localization model.
The representative THUMOS14~\cite{THUMOS14} is selected as the test data for its high ratio between the background and foreground. 
From \cref{tab:tal_comp}, our method achieves 44.5$\%$ in mAP@0.5, a 1.3$\%$ gain over the TSP~\cite{Alwassel2021TSPTP}. By using our extracted video features, the GTAD observes significant improvement.

\subsubsection{Video Moment Retrieval}
We also evaluate our method on the moment retrieval task where the temporal boundary in a video is predicted for the text description. We follow LocVTP~\cite{Cao2022LocVTPVP} to use the 2D-TAN~\cite{2DTAN_2020_AAAI} as the baseline model for comparison. 
\cref{tab:moment_ret_comp} provides the video moment retrieval results on DiDeMo. 
Although the same pre-training datasets are used, our method outperforms LocVTP by a large margin (4.3) on the Rank@0.5. It shows that temporal boundaries are more accurate in the top retrieval results with our extracted features.

\begin{table}
  \centering
  \resizebox{0.45\textwidth}{!}{
  \begin{tabular}{l|cccc}
    \toprule
    Method & mAP@0.3 & mAP@0.5 & mAP@0.7 & AVG \\
    \midrule  
    GTAD~\cite{xu2020gtad} & 57.3 & 43.0 & 22.8 & 41.0 \\
    TAL-Net~\cite{Chao2018RethinkingTF} & 53.2 & 42.8 & 20.8 & 38.9  \\ 
    TSP~\cite{Alwassel2021TSPTP} & \textbf{59.6} & 43.2 & 21.1 & 41.3 \\
    \midrule
    TemPVL (Ours) & 58.9 & \textbf{44.5} & \textbf{24.1} & \textbf{42.5} \\
    \bottomrule
  \end{tabular}
  }
  \caption{Temporal action localization comparison on THUMOS14. The mAP@0.3 denotes the mean average precision at temporal IoU 0.3 and AVG is the average score of the three metrics. }
  \label{tab:tal_comp}
\end{table}

\begin{table}
  \centering
  \begin{tabular}{l|ccc}
    \toprule
    Method & Rank@0.5 & Rank@0.7 & AVG \\
    \midrule 
    2D-TAN~\cite{2DTAN_2020_AAAI} & 42.8 & 23.2 & 33.0 \\
    LocVTP~\cite{Cao2022LocVTPVP} & 41.2 & \textbf{24.8} & 33.0 \\
    \midrule
    TemPVL (Ours) & \textbf{45.4} & 24.7 & \textbf{35.1} \\
    \bottomrule
  \end{tabular}
  \caption{Moment retrieval comparison on Charades-STA. The Rank@0.5 denotes the top-1 retrieval results with temporal IoU greater than 0.5. AVG indicates the average score of two metrics.}
  \label{tab:moment_ret_comp}
\end{table}

\subsection{Analysis}
We conduct ablation experiments on WebVid1M which is a subset of WebVid dataset containing one million video-text pairs, to study the effectiveness of the proposed designs.

\subsubsection{Pre-training Objective}
As shown in \cref{tab:pre_task}, we evaluate our proposed objectives on three tasks, including text-to-video retrieval, video question answering and captioning.  Compared to the baseline model which uses contrastive learning and masked language modeling, our model with two localization objectives significantly improves the zero-shot retrieval performance by 1.6$\%$ on recall@1. 
The model pre-trained with moment retrieval task acquires highest accuracy on VQA and obtains higher performance gain than the model pre-trained with the text localization objective.
For the video captioning task, the improvement is most significant when the model pre-trained with both objectives.
On all three tasks, our model pre-trained with text-video localization outperforms the baseline model. This suggests that the model pre-trained with the text-video localization task obtains superior 
generalization capacity.

\subsubsection{Video Merging}
For the moment localization with the language query, we adopt three strategies in ~\cref{tab:vis_strategy} to combine individual video features into a long video sequence. The quality of the merged sequence is essential since the localization task is to predict the temporal boundary from the sequence. The Shuffling strategy only changes the order of different videos where the order of frame features is preserved. Sampling strategy requires the max length to specify how many frames should be sampled. The positive number of frames is also required to define how many frames are related to the text descriptions.
In this experiment, we specify the max sample number $K$ to 128, and the minimum and the maximum number of positive frames $K_{p}$ to 1 and 32. 
The HardSampling denotes that frame features are only sampled from those most similar videos in a batch.  Specifically, we calculate all the similarities between video features. Frame feature is then sampled from the top 10 videos with the highest similarity score. 
From~\cref{tab:vis_strategy}, we observe that the HardSampling strategy acquires the highest zero-shot retrieval performance. It shows that merging frame tokens from similar videos for moment localization is more helpful to the retrieval task.

\begin{table}[t]
\centering
	\begin{tabular}{l|ccc}
	\toprule
		            & Retrieval  & VQA  & Captioning \\
	\midrule
		Baseline &  20.8 & 42.7 & 56.9 \\
		+$\mathcal{L}_{vl}$ & 22.1 & \textbf{43.1} & 57.7  \\
		+$\mathcal{L}_{tl}$ & 21.9 & 42.7 & 57.5 \\
		+$\mathcal{L}_{vtl}$ & \textbf{22.4} & 43.0 & \textbf{58.1} \\
	\bottomrule
	\end{tabular}
	\caption{Effect of pretraining tasks on downstream tasks. The recall@1, accuracy, and CIDEr are reported in the zero-shot text-to-video retrieval, video question answering, and video captioning tasks, respectively.}
	\label{tab:pre_task}
\end{table}

\begin{table}[t]

\centering
	\begin{tabular}{l|cccc}
	\toprule
		Strategy            & Retrieval  & VQA  & Captioning \\
	\midrule
		Shuffling & 21.7 & 42.9 & \textbf{58.1} \\
		Sampling & 21.5 & 42.8 & 56.9 \\
		HardSampling & \textbf{22.1} & \textbf{43.1} & 57.7 \\
	\bottomrule
	\end{tabular}
	\caption{Analysis of video merging strategies for moment retrieval pre-training task. Results on text-to-video retrieval, video question answering, and video captioning tasks are reported.}
	\label{tab:vis_strategy}
\end{table}

\begin{figure*}
  \centering
  \includegraphics[width=\linewidth]{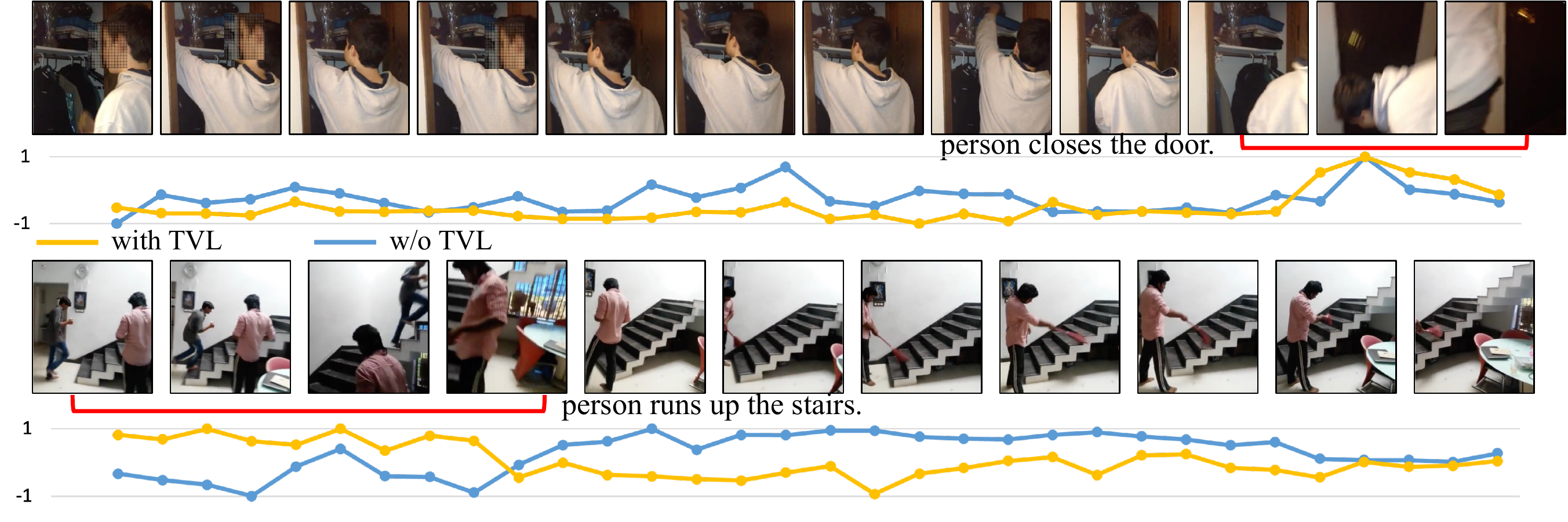}
    \caption{Visualization of similarity between frame and text features. The temporal ground-truth for the text description is marked with red.}
    \label{fig:visual}
  
\end{figure*}

\subsubsection{Text Merging}
For the text localization task, we compare two merging strategies in~\cref{tab:text_strategy}. 
The word merging is similar to the shuffling in video merging, where only the sentence order is rearranged. 
We retain the special tokens ([CLS] and [SEP]) when merging words. We predict the start and end positions for the video in this merging strategy. 
CLS merging denotes that only the first token in each sentence is selected to combine different texts.
We found that the model with CLS merging achieves a bit better retrieval performance from the experimental results. 
Since only the [CLS] token feature is used in the text-to-video retrieval task, using the [CLS] token for the text localization is beneficial to align the text representations with different video features.

\subsubsection{Effect of Visual Backbones}
We summarize the experimental results of our method with different backbones on five tasks in \cref{tab:backbone}. Two visual backbones, including a video encoder SwinT~\cite{Liu2022VideoST} and a frame encoder ViT~\cite{Dosovitskiy2021AnII}, are adopted to verify the text-video localization pre-training. SwinT and ViT are initialized with weights pre-trained on Kinetics and ImageNet, respectively. On MSR-VTT, the model with TVL obtains significant improvement in the zero-shot text-to-video retrieval task for both video and image encoders. For the temporal action localization on THUMOS14, we only extract features on RGB frames with the pre-trained model and use the GTAD~\cite{xu2020gtad} for training. Both ViT and SwinT based models with TVL pre-training obtain about 1$\%$ performance gain on the mAP@0.5 metric. Our method also significantly improves the Rank@0.5 on Charades. 
Albeit ViT based model achieves lower performance than the SwinT based model, both visual encoders with the TVL pre-training obtain improvement on almost all tasks.
This demonstrates that generic representations are well learned with our proposed text-video localization task.

\begin{table}[t]
\centering
	\begin{tabular}{l|ccc}
	\toprule
		Strategy            & Retrieval  & VQA  & Captioning \\
	\midrule
		Merge words & 21.7  & \textbf{42.9} & 57.4 \\
		Merge CLS &  \textbf{21.9} & 42.7 & \textbf{57.5} \\
	\bottomrule
	\end{tabular}
	\caption{Analysis of word merging strategies for text localization pre-training task. Results on text-to-video retrieval, video question answering, and video captioning tasks are reported.}
	\label{tab:text_strategy}
\end{table}

\subsubsection{Qualitative Examples}
We further analyze how the model pre-trained with the text-video localization task performs better on the downstream tasks. 
We extract frame and text features on the Charades-STA~\cite{anne2017localizing}, and calculate the similarity between the visual and language features. 
Each video is about 30s and contains several actions and events.
We visualize the results in \cref{fig:visual} to show the difference between models trained with and without text-video localization. 
For the video with caption ``person runs up the stairs'', although the global match scores of the two models are close, the model only with TVL pre-training accurately localizes frames that match the caption.
It shows that the cross modality features are better aligned in our method, and the temporal boundaries are more accurate for text descriptions.

\begin{table}[t]
\centering
\resizebox{\linewidth}{!}{
	\begin{tabular}{l|ccc|c|c}
	\toprule
	   \multicolumn{1}{l|}{\multirow{2}{*}{Backbone}} & \multicolumn{3}{c|}{MSR-VTT} & THUMOS & Charades \\
		          & Recall@1  & Acc  & CIDEr & mAP@0.5 & Rank@0.5  \\
	\midrule
	
		ViT~(w/o TVL)  & 16.4 & 36.2 & 53.1 & 20.2 & 38.8 \\
		ViT~(w TVL)  &  17.7 & 36.9 & 53.0 & 20.8 &40.2 \\ \midrule
		SwinT~(w/o TVL)  & 20.8 & 42.7 & 56.9 & 29.5 & 41.8 \\
		SwinT~(w TVL)  & 22.4 & 43.1 & 58.1 & 30.6 & 43.6 \\ 
	\bottomrule
	\end{tabular}}
	\caption{Effect of text-video localization pre-training task with different visual backbones. Results on three datasets with five tasks are reported.}
	\label{tab:backbone}
\end{table}

\section{Conclusion}
\label{sec:conc}

We introduce a text-video localization task for video-language pre-training. Without any temporal annotations, we construct long sequences by merging video and text features and use the multi-modal encoder to predict the boundary. The proposed localization task is simple yet effective to learn a generic muliti-modal representations. Extensive experiments on text-video retrieval, temporal action localization and  moment localization tasks also demonstrate the strength of the proposed pre-text task. 

\appendix

\section{Experimental Results}

\begin{table}
\centering
	\begin{tabular}{l|cccc}
	\toprule
		Strategy            & Retrieval  & VQA  & Captioning \\
	\midrule
		MM3 & 22.4 & 43.0 & 58.1 \\
		MM6 & 21.7 & 43.1 & 58.3 \\
		MM12 & 20.9 & 42.7 & 58.3 \\
	\bottomrule
	\end{tabular}
	\caption{Impact of Multi-modal encoder on text-to-video retrieval, video question answering, and video captioning tasks. MM3 denotes that 3 Bert layers are used in the Multi-modal encoder.}
	\label{tab:mm_layer}
\end{table}

\subsection{Multi-modal Encoder}
For the multi-modal encoder, we use a pre-trained Bert model to initialize it. In this section, we study the impact of multi-modal encoder on different tasks in \cref{tab:mm_layer}. Specifically, we use the top layer of the Bert where the MM3 denotes that the last 3 layers of Bert are used. Since the multi-modal encoder is not needed in the text-to-video retrieval task, adding more layers seems not helpful for the global video text alignment. For video captioning which requires the multi-modal encoder, higher results are obtained when more layers are adopted. 
For VQA, the highest accuracy is observed when six Bert layers are used. 

\subsection{Video Sampling}
We also study the sampling strategy for merging different videos in \cref{tab:sampling}.  It shows that the model achieves inferior performance when the merged sequence is not long. For the captioning task, the positive frames are significant during pre-training. The model performs better when the maximum number of positive frames is set to a large value. Note there could be less than $K_{p}$ positive frames since we randomly sample $k$ positive frames during merging where $k \leq K_{p}$. 

\begin{table}
\centering
	\begin{tabular}{ll|cccc}
	\toprule
		$K$ & $K_{p}$          & Retrieval  & VQA  & Captioning \\
	\midrule
	    32 & 16 & 21.5 & 42.8 & 57.5 \\
	    128 & 16 & 21.4 & 43.0 & 57.3  \\
		128 & 32 &  22.1 & 43.0 & 58.2 \\
	\bottomrule
	\end{tabular}
	\caption{Analysis of sampling merging on text-to-video retrieval, video question answering, and video captioning tasks. $K$ denotes the maximum number of total frames in the merged videos and $K_p$ denotes the maximum number of positive frames that contain related frames for the merged text.}
	\label{tab:sampling}
\end{table}

\begin{figure}[!ht]
    \includegraphics[width=\linewidth]{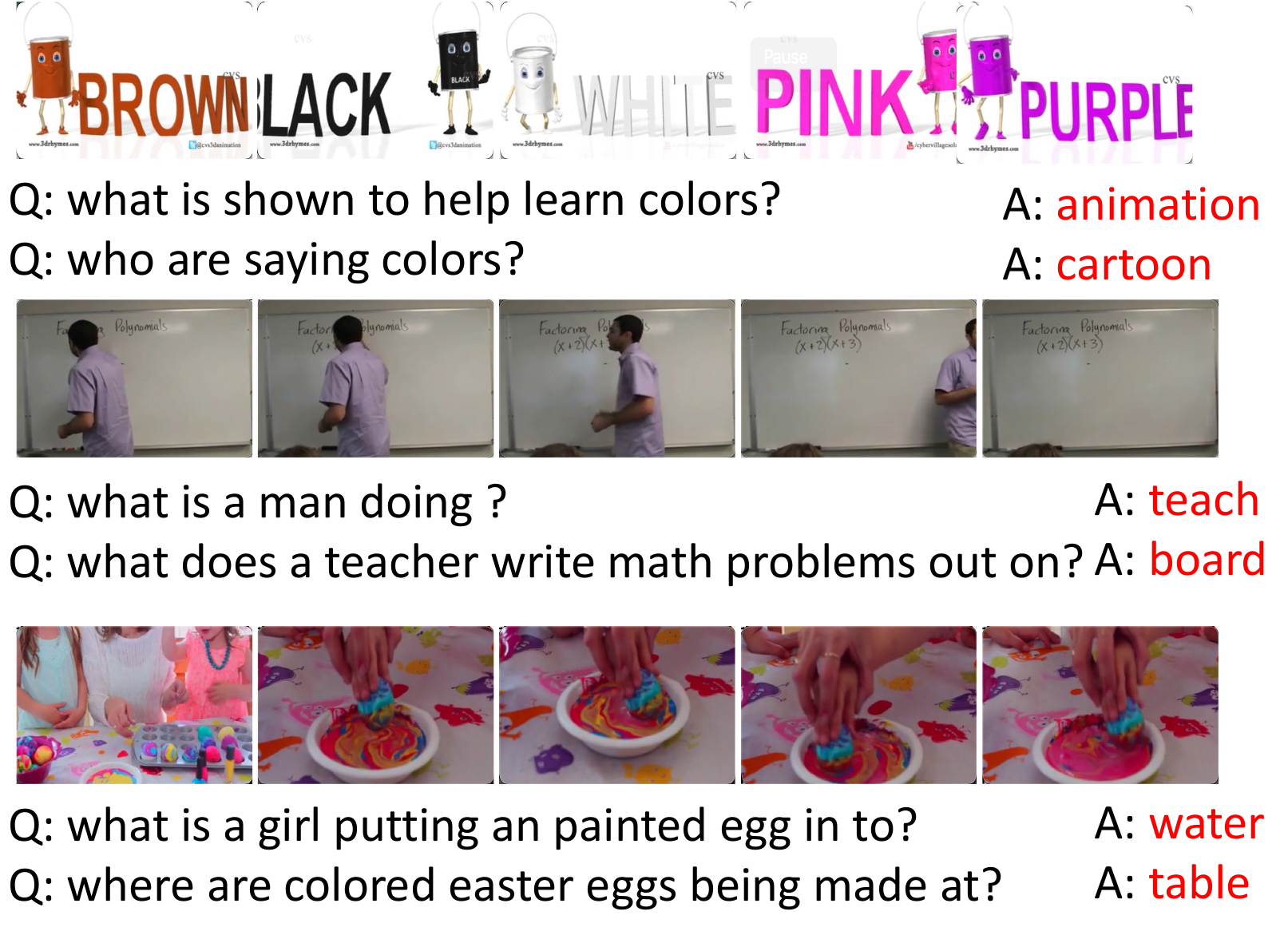}
    \caption{Visualization of the predictions on the video question answering. The videos are from MSR-VTT. 
    }
    \label{fig:vqa} 
\end{figure}

\subsection{Video Question Answering}
We show predicted answers for some video questions on MSR-VTT in \cref{fig:vqa}. 
For different types of videos, our model predicts the correct answer even though our model does not use the classification for open-ended question answering. 
This demonstrates that our model performs well without limiting the range of answers.  

\begin{figure}[!ht]
    \includegraphics[width=\linewidth]{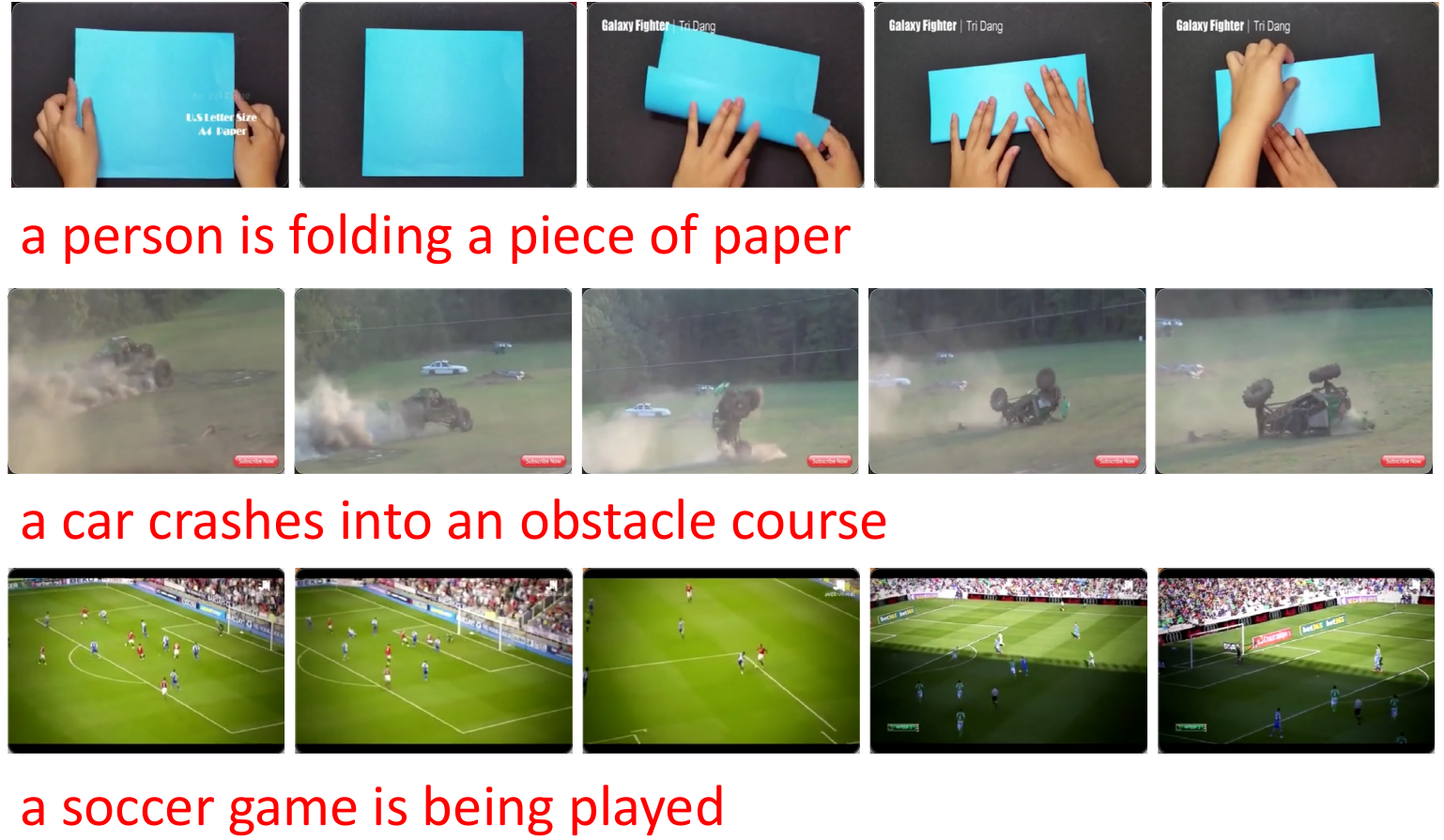}
    \caption{Generated captions on the MSR-VTT videos. 
    }
    \label{fig:caption} 
\end{figure}

\subsection{Video Captionining}
We show the generated captions on MSR-VTT in \cref{fig:caption}. It is shown that our model clearly identifies the object and scenes of videos. Although most of the scenes are recognized by our method, the fine-grained details in videos such as the ``score moment'' is not present in the caption. In the future, we will consider how to generate more details for video captioning to include more fine-grained actions and moment descriptions.

{\small
\bibliographystyle{ieee_fullname}
\bibliography{egbib}

\begin{thebibliography}{10}\itemsep=-1pt

\bibitem{Alwassel2021TSPTP}
Humam Alwassel, Silvio Giancola, and Bernard Ghanem.
\newblock Tsp: Temporally-sensitive pretraining of video encoders for
  localization tasks.
\newblock {\em 2021 IEEE/CVF International Conference on Computer Vision
  Workshops (ICCVW)}, pages 3166--3176, 2021.

\bibitem{amrani2021noise}
Elad Amrani, Rami Ben-Ari, Daniel Rotman, and Alex Bronstein.
\newblock Noise estimation using density estimation for self-supervised
  multimodal learning.
\newblock In {\em Proceedings of the AAAI Conference on Artificial
  Intelligence}, volume~35, pages 6644--6652, 2021.

\bibitem{anne2017localizing}
Lisa Anne~Hendricks, Oliver Wang, Eli Shechtman, Josef Sivic, Trevor Darrell,
  and Bryan Russell.
\newblock Localizing moments in video with natural language.
\newblock In {\em Proceedings of the IEEE international conference on computer
  vision}, pages 5803--5812, 2017.

\bibitem{Bain21}
Max Bain, Arsha Nagrani, G{\"u}l Varol, and Andrew Zisserman.
\newblock Frozen in time: A joint video and image encoder for end-to-end
  retrieval.
\newblock In {\em IEEE International Conference on Computer Vision}, 2021.

\bibitem{bain2021frozen}
Max Bain, Arsha Nagrani, G{\"u}l Varol, and Andrew Zisserman.
\newblock Frozen in time: A joint video and image encoder for end-to-end
  retrieval.
\newblock In {\em Proceedings of the IEEE/CVF International Conference on
  Computer Vision}, pages 1728--1738, 2021.

\bibitem{bertasius2021space}
Gedas Bertasius, Heng Wang, and Lorenzo Torresani.
\newblock Is space-time attention all you need for video understanding?
\newblock In {\em ICML}, volume~2, page~4, 2021.

\bibitem{Cao2022LocVTPVP}
Meng Cao, Tianyu Yang, Junwu Weng, Can Zhang, Jue Wang, and Yuexian Zou.
\newblock Locvtp: Video-text pre-training for temporal localization.
\newblock {\em ArXiv}, abs/2207.10362, 2022.

\bibitem{Chao2018RethinkingTF}
Yu-Wei Chao, Sudheendra Vijayanarasimhan, Bryan Seybold, David~A. Ross, Jia
  Deng, and Rahul Sukthankar.
\newblock Rethinking the faster r-cnn architecture for temporal action
  localization.
\newblock {\em 2018 IEEE/CVF Conference on Computer Vision and Pattern
  Recognition}, pages 1130--1139, 2018.

\bibitem{Dosovitskiy2021AnII}
Alexey Dosovitskiy, Lucas Beyer, Alexander Kolesnikov, Dirk Weissenborn,
  Xiaohua Zhai, Thomas Unterthiner, Mostafa Dehghani, Matthias Minderer, Georg
  Heigold, Sylvain Gelly, Jakob Uszkoreit, and Neil Houlsby.
\newblock An image is worth 16x16 words: Transformers for image recognition at
  scale.
\newblock {\em ArXiv}, abs/2010.11929, 2021.

\bibitem{feichtenhofer2019slowfast}
Christoph Feichtenhofer, Haoqi Fan, Jitendra Malik, and Kaiming He.
\newblock Slowfast networks for video recognition.
\newblock In {\em Proceedings of the IEEE/CVF international conference on
  computer vision}, pages 6202--6211, 2019.

\bibitem{Fu2021VIOLETE}
Tsu-Jui Fu, Linjie Li, Zhe Gan, Kevin Lin, William~Yang Wang, Lijuan Wang, and
  Zicheng Liu.
\newblock Violet : End-to-end video-language transformers with masked
  visual-token modeling.
\newblock {\em ArXiv}, abs/2111.12681, 2021.

\bibitem{ge2022bridging}
Yuying Ge, Yixiao Ge, Xihui Liu, Dian Li, Ying Shan, Xiaohu Qie, and Ping Luo.
\newblock Bridging video-text retrieval with multiple choice questions.
\newblock In {\em Proceedings of the IEEE/CVF Conference on Computer Vision and
  Pattern Recognition}, pages 16167--16176, 2022.

\bibitem{huang2022clover}
Jingjia Huang, Yinan Li, Jiashi Feng, Xiaoshuai Sun, and Rongrong Ji.
\newblock Clover: Towards a unified video-language alignment and fusion model.
\newblock {\em arXiv preprint arXiv:2207.07885}, 2022.

\bibitem{THUMOS14}
Y.-G. Jiang, J. Liu, A. Roshan~Zamir, G. Toderici, I. Laptev, M. Shah, and R.
  Sukthankar.
\newblock {THUMOS} challenge: Action recognition with a large number of
  classes.
\newblock \url{http://crcv.ucf.edu/THUMOS14/}, 2014.

\bibitem{Kay2017TheKH}
Will Kay, Jo{\~a}o Carreira, Karen Simonyan, Brian Zhang, Chloe Hillier,
  Sudheendra Vijayanarasimhan, Fabio Viola, Tim Green, Trevor Back, Apostol
  Natsev, Mustafa Suleyman, and Andrew Zisserman.
\newblock The kinetics human action video dataset.
\newblock {\em ArXiv}, abs/1705.06950, 2017.

\bibitem{clipbert2020}
Jie Lei, Linjie Li, Luowei Zhou, Zhe Gan, Tamara~L Berg, Mohit Bansal, and
  Jingjing Liu.
\newblock Less is more: Clipbert for video-and-language learning via sparse
  sampling.
\newblock In {\em Proceedings of the IEEE/CVF Conference on Computer Vision and
  Pattern Recognition}, pages 7331--7341, 2021.

\bibitem{Li2022AlignAP}
Dongxu Li, Junnan Li, Hongdong Li, Juan~Carlos Niebles, and Steven C.~H. Hoi.
\newblock Align and prompt: Video-and-language pre-training with entity
  prompts.
\newblock pages 4943--4953, 2022.

\bibitem{li2020hero}
Linjie Li, Yen-Chun Chen, Yu Cheng, Zhe Gan, Licheng Yu, and Jingjing Liu.
\newblock Hero: Hierarchical encoder for video+ language omni-representation
  pre-training.
\newblock {\em arXiv preprint arXiv:2005.00200}, 2020.

\bibitem{li2022lavender}
Linjie Li, Zhe Gan, Kevin Lin, Chung-Ching Lin, Zicheng Liu, Ce Liu, and Lijuan
  Wang.
\newblock Lavender: Unifying video-language understanding as masked language
  modeling.
\newblock {\em arXiv preprint arXiv:2206.07160}, 2022.

\bibitem{Lin2022SwinBERTET}
Kevin Lin, Linjie Li, Chung-Ching Lin, Faisal Ahmed, Zhe Gan, Zicheng Liu,
  Yumao Lu, and Lijuan Wang.
\newblock Swinbert: End-to-end transformers with sparse attention for video
  captioning.
\newblock pages 17928--17937, 2022.

\bibitem{Liu2022VideoST}
Ze Liu, Jia Ning, Yue Cao, Yixuan Wei, Zheng Zhang, Stephen Lin, and Han Hu.
\newblock Video swin transformer.
\newblock {\em 2022 IEEE/CVF Conference on Computer Vision and Pattern
  Recognition (CVPR)}, pages 3192--3201, 2022.

\bibitem{Loshchilov2019DecoupledWD}
Ilya Loshchilov and Frank Hutter.
\newblock Decoupled weight decay regularization.
\newblock In {\em ICLR}, 2019.

\bibitem{maharaj2017dataset}
Tegan Maharaj, Nicolas Ballas, Anna Rohrbach, Aaron Courville, and Christopher
  Pal.
\newblock A dataset and exploration of models for understanding video data
  through fill-in-the-blank question-answering.
\newblock In {\em Proceedings of the IEEE Conference on Computer Vision and
  Pattern Recognition}, pages 6884--6893, 2017.

\bibitem{miech2020end}
Antoine Miech, Jean-Baptiste Alayrac, Lucas Smaira, Ivan Laptev, Josef Sivic,
  and Andrew Zisserman.
\newblock End-to-end learning of visual representations from uncurated
  instructional videos.
\newblock In {\em Proceedings of the IEEE/CVF Conference on Computer Vision and
  Pattern Recognition}, pages 9879--9889, 2020.

\bibitem{miech19howto100m}
Antoine Miech, Dimitri Zhukov, Jean-Baptiste Alayrac, Makarand Tapaswi, Ivan
  Laptev, and Josef Sivic.
\newblock How{T}o100{M}: {L}earning a {T}ext-{V}ideo {E}mbedding by {W}atching
  {H}undred {M}illion {N}arrated {V}ideo {C}lips.
\newblock In {\em ICCV}, 2019.

\bibitem{patrick2020support}
Mandela Patrick, Po-Yao Huang, Yuki Asano, Florian Metze, Alexander Hauptmann,
  Joao Henriques, and Andrea Vedaldi.
\newblock Support-set bottlenecks for video-text representation learning.
\newblock In {\em ICLR}, 2020.

\bibitem{clip2021}
Alec Radford, Jong~Wook Kim, Chris Hallacy, Aditya Ramesh, Gabriel Goh,
  Sandhini Agarwal, Girish Sastry, Amanda Askell, Pamela Mishkin, Jack Clark,
  Gretchen Krueger, and Ilya Sutskever.
\newblock Learning transferable visual models from natural language
  supervision.
\newblock In Marina Meila and Tong Zhang, editors, {\em Proceedings of the 38th
  International Conference on Machine Learning}, volume 139 of {\em Proceedings
  of Machine Learning Research}, pages 8748--8763. PMLR, 18--24 Jul 2021.

\bibitem{Seo2022EndtoendGP}
Paul~Hongsuck Seo, Arsha Nagrani, Anurag Arnab, and Cordelia Schmid.
\newblock End-to-end generative pretraining for multimodal video captioning.
\newblock pages 17938--17947, 2022.

\bibitem{Sharma2018ConceptualCA}
Piyush Sharma, Nan Ding, Sebastian Goodman, and Radu Soricut.
\newblock Conceptual captions: A cleaned, hypernymed, image alt-text dataset
  for automatic image captioning.
\newblock In {\em ACL}, 2018.

\bibitem{sun2019videobert}
Chen Sun, Austin Myers, Carl Vondrick, Kevin Murphy, and Cordelia Schmid.
\newblock Videobert: A joint model for video and language representation
  learning.
\newblock In {\em Proceedings of the IEEE/CVF International Conference on
  Computer Vision}, pages 7464--7473, 2019.

\bibitem{tang2021decembert}
Zineng Tang, Jie Lei, and Mohit Bansal.
\newblock Decembert: Learning from noisy instructional videos via dense
  captions and entropy minimization.
\newblock In {\em Proceedings of the 2021 Conference of the North American
  Chapter of the Association for Computational Linguistics: Human Language
  Technologies}, pages 2415--2426, 2021.

\bibitem{tran2014c3d}
Du Tran, Lubomir~D Bourdev, Rob Fergus, Lorenzo Torresani, and Manohar Paluri.
\newblock C3d: generic features for video analysis.
\newblock {\em CoRR, abs/1412.0767}, 2(7):8, 2014.

\bibitem{wang2022all}
Alex~Jinpeng Wang, Yixiao Ge, Rui Yan, Yuying Ge, Xudong Lin, Guanyu Cai,
  Jianping Wu, Ying Shan, Xiaohu Qie, and Mike~Zheng Shou.
\newblock All in one: Exploring unified video-language pre-training.
\newblock {\em arXiv preprint arXiv:2203.07303}, 2022.

\bibitem{TSN2016ECCV}
Limin Wang, Yuanjun Xiong, Zhe Wang, Yu Qiao, Dahua Lin, Xiaoou Tang, and Luc
  {Val Gool}.
\newblock Temporal segment networks: Towards good practices for deep action
  recognition.
\newblock In {\em ECCV}, 2016.

\bibitem{xu2017video}
Dejing Xu, Zhou Zhao, Jun Xiao, Fei Wu, Hanwang Zhang, Xiangnan He, and Yueting
  Zhuang.
\newblock Video question answering via gradually refined attention over
  appearance and motion.
\newblock In {\em Proceedings of the 25th ACM international conference on
  Multimedia}, pages 1645--1653, 2017.

\bibitem{xu2021videoclip}
Hu Xu, Gargi Ghosh, Po-Yao Huang, Dmytro Okhonko, Armen Aghajanyan, Florian
  Metze, Luke Zettlemoyer, and Christoph Feichtenhofer.
\newblock Videoclip: Contrastive pre-training for zero-shot video-text
  understanding.
\newblock {\em arXiv preprint arXiv:2109.14084}, 2021.

\bibitem{xu2016msr}
Jun Xu, Tao Mei, Ting Yao, and Yong Rui.
\newblock Msr-vtt: A large video description dataset for bridging video and
  language.
\newblock In {\em Proceedings of the IEEE conference on computer vision and
  pattern recognition}, pages 5288--5296, 2016.

\bibitem{xu2020gtad}
Mengmeng Xu, Chen Zhao, David~S. Rojas, Ali Thabet, and Bernard Ghanem.
\newblock G-tad: Sub-graph localization for temporal action detection.
\newblock In {\em Proceedings of the IEEE/CVF Conference on Computer Vision and
  Pattern Recognition (CVPR)}, 2020.

\bibitem{Yang2021JustAL}
Antoine Yang, Antoine Miech, Josef Sivic, Ivan Laptev, and Cordelia Schmid.
\newblock Just ask: Learning to answer questions from millions of narrated
  videos.
\newblock pages 1666--1677, 2021.

\bibitem{zellers2021merlot}
Rowan Zellers, Ximing Lu, Jack Hessel, Youngjae Yu, Jae~Sung Park, Jize Cao,
  Ali Farhadi, and Yejin Choi.
\newblock Merlot: Multimodal neural script knowledge models.
\newblock {\em Advances in Neural Information Processing Systems},
  34:23634--23651, 2021.

\bibitem{Zellers2021MERLOTMN}
Rowan Zellers, Ximing Lu, Jack Hessel, Youngjae Yu, Jae~Sung Park, Jize Cao,
  Ali Farhadi, and Yejin Choi.
\newblock Merlot: Multimodal neural script knowledge models.
\newblock In {\em NeurIPS}, 2021.

\bibitem{2DTAN_2020_AAAI}
Songyang Zhang, Houwen Peng, Jianlong Fu, and Jiebo Luo.
\newblock Learning 2d temporal adjacent networks formoment localization with
  natural language.
\newblock In {\em AAAI}, 2020.

\bibitem{zhu2020actbert}
Linchao Zhu and Yi Yang.
\newblock Actbert: Learning global-local video-text representations.
\newblock In {\em Proceedings of the IEEE/CVF conference on computer vision and
  pattern recognition}, pages 8746--8755, 2020.

\end{thebibliography}
}

\end{document}